\documentclass[10pt,a4paper]{article}
\usepackage{graphicx}
\usepackage{multirow}
\usepackage{amsmath}
\usepackage{amssymb}
\usepackage{mathtools}
\usepackage{booktabs}
\usepackage{algorithm}
\usepackage{algorithmic}
\usepackage{tikz}
\usepackage{array}
\usepackage{tabularx}
\usepackage{xcolor}
\usepackage{caption}
\usepackage{subcaption}
\usepackage{authblk} % Package for author affiliations
\usepackage{geometry}
\usepackage{hyperref}

\usepackage{times}
\usepackage{newtxtext,newtxmath}

\usepackage{setspace}
\begin{document}

\title{\Large\textbf{A Hybrid DeBERTa and Gated Broad Learning System for Cyberbullying Detection in English Text}}

\author[1]{Devesh Kumar\thanks{23mcs112@nith.ac.in}}
\affil[1]{Department of Computer Science \& Engineering, National Institute of Technology Hamirpur, Hamirpur (H.P.) - 177005, India}

\date{}  % No date

\maketitle

\begin{abstract}
The proliferation of online communication platforms has created unprecedented opportunities for global connectivity while simultaneously enabling harmful behaviors such as cyberbullying, which affects approximately 54.4\% of teenagers according to recent research. This paper presents a hybrid architecture that combines the contextual understanding capabilities of transformer-based models with the pattern recognition strengths of broad learning systems for effective cyberbullying detection. This approach integrates a modified DeBERTa model augmented with Squeeze-and-Excitation blocks and sentiment analysis capabilities with a Gated Broad Learning System (GBLS) classifier, creating a synergistic framework that outperforms existing approaches across multiple benchmark datasets. The proposed ModifiedDeBERTa + GBLS model achieved good performance on four English datasets: 79.3\% accuracy on HateXplain, 95.41\% accuracy on SOSNet, 91.37\% accuracy on Mendeley-I, and 94.67\% accuracy on Mendeley-II. Beyond performance gains, the framework incorporates comprehensive explainability mechanisms including token-level attribution analysis, LIME-based local interpretations, and confidence calibration, addressing critical transparency requirements in automated content moderation. Ablation studies confirm the meaningful contribution of each architectural component, while failure case analysis reveals specific challenges in detecting implicit bias and sarcastic content, providing valuable insights for future improvements in cyberbullying detection systems.
\end{abstract}


% Body text starts here
\section{Introduction}

The digital landscape has been transformed through rapid growth of online communication platforms, enabling unprecedented global connectivity while simultaneously creating new avenues for harmful behaviors. Social networks have fundamentally changed how people interact, but have also facilitated troubling behaviors such as cyberbullying—a destructive form of online harassment that causes psychological damage to victims and undermines healthy community interactions \cite{kowalski2014psychological}. Unlike traditional bullying, cyberbullying continues indefinitely, reaches victims immediately regardless of physical distance, and leaves permanent digital traces with potentially lasting harm.

Cyberbullying refers to the deliberate and repeated use of digital technologies to harass, threaten, humiliate, or harm others. It is characterized by an intent to harm, repetitive patterns, power imbalances, and personal targeting. These characteristics distinguish cyberbullying from valid online emotions such as spirited arguments, positive criticism, or unpopular opinions. Recent studies at the Cyberbullying Research Center have determined that nearly 54.4\% of teenagers have been victims of cyberbullying, and approximately 26.4\% have reported incidents over a 30-day timeline \cite{patchin2023cyberbullying}. Such chilling figures highlight the importance of effective methods of detection.

This research addresses these challenges through the development of a hybrid architecture that combines the contextual understanding capabilities of transformer-based models with the pattern recognition strengths of broad learning systems. The proposed framework integrates a modified DeBERTa model with a Gated Broad Learning System (GBLS) classifier, augmented by sentiment analysis and feature selection mechanisms to achieve better classification performance while maintaining interpretability.

The main contributions of this work are: (1) the creation of a hybrid DeBERTa-GBLS solution that outperforms all other methods on a number of benchmark sets across the board, (2) designing an end-to-end explainability module with token-level attributions, confidence calibration, and LIME explanations, and (3) thorough experimental evaluation through ablation studies and failure case analysis for verifying model robustness and limitation.

The remainder of this paper is organized as follows: Section 2 surveys existing cyberbullying detection techniques in the literature, examining baseline models, advanced transformer architectures, and hybrid approaches. Section 3 describes the proposed methodological framework, including the ModifiedDeBERTa component, GBLS classifier. Section 4 details the experimental setup, including datasets, implementation details, and evaluation metrics. Section 5 presents comprehensive evaluation results and interpretability analyses. Section 6 discusses the explainability framework and its implications for understanding model decisions compared to existing approaches. Finally, Section 7 concludes the paper and outlines promising directions for future research.

\section{Related Work}

This section reviews existing approaches to cyberbullying detection, examining the evolution from traditional machine learning methods to more sophisticated deep learning and transformer-based architectures.

The earliest cyberbullying detection systems were primarily based on traditional machine learning approaches built on hand-crafted features. Schmidt and Wiegand \cite{schmidt2017} present an extensive overview of natural language processing techniques used for the purpose of hate speech detection, encompassing the transition from traditional machine learning to more recent trends. The traditional systems relied on lexical features like the usage of swear words and applied statistical classification approaches like Support Vector Machines (SVM) and Naive Bayes classifiers.

Davidson et al. \cite{davidson2017} addressed the issue of distinguishing between real hate speech and offending non-hate speech by developing a three-way classification Twitter corpus. They used character n-grams, word n-grams, and word skip-grams as features and achieved 78\% accuracy with their system. Their work placed significance on the necessity of distinguishing between classification that determines a variety of degrees of harmful intent rather than mere binary classification.

While historically useful, traditional approaches are poor at capturing contextual nuance and evolving linguistic patterns characteristic of cyberbullying discourse. Relying on stable lexicocanonical features is likely to generate high rates of false positives, particularly in detecting insidiously filthy obscenity irrespective of overtly degrading talk.

The advent of deep learning architectures was a major step forward in cyberbullying detection performance. Badjatiya et al. \cite{badjatiya2017} performed systematic tests on several deep learning structures to create better lexical representations. On a corpus of 16,000 labeled tweets, they showed the excellent performance of deep learning methods over conventional approaches with F1-score gains of around 18 percentage points. Their best results came from hybrid methods: applying deep neural networks to generate dense word embeddings, and then using these as input features with gradient-boosted decision trees for classification.

The capabilities of recurrent neural networks, especially those that utilize gating mechanisms, have remarkably modified sequence processing for text. First introduced in Long Short-Term Memory (LSTM) networks by Hochreiter and Schmidhuber \cite{hochreiter1997long}, intelligent filters that selectively control information flow conditionally could be utilized. GRUs, in turn, were developed later by Cho et al. \cite{cho2014learning} with the elimination of some gating operations in LSTMs, such that the performance results are on par. This simplification demonstrated that efficient information control does not necessarily require complex architectural design, prompting exploration into novel gated neural methods with increasingly refined gating integration techniques across diverse model architectures.

Deep learning techniques significantly capture sequential dependencies and semantic relationships in the text itself. This allows better classification than the traditional methods. However, these techniques typically require a lot of training data and computational resources and lack interpretability due to their black-box nature. The recent transformations in architectures have also upgraded natural language processing applications such as cyberbullying detection to some extent, overcoming several of these challenges. Many specialized transformer models have been proven quite effective for the detection of toxic content, with better performance while still struggling to overcome problems of explainability and resource-hungry characteristics.

Developed on top of the baseline BERT model, Caselli et al. \cite{caselli2021hatebert} enhance it with targeted pre-training on abusive data web-scraped from Reddit forums. Such targeted tuning allows HateBERT to better detect linguistic patterns in hate speech and cyberbullying. Targeted training improves the capacity for the model to identify linguistic signs of ill intent. HateBERT is, therefore, a highly recommendable benchmark to analyze models that attack cyberbullying.

Barbieri et al. \cite{barbieri2020tweeteval} created XLM-T, a transformer-based model based on the XLM-RoBERTa model but with a unique pre-training on Twitter datasets in multiple languages. Its emphasis on social media language makes XLM-T especially well-suited to the identification of cyberbullying in informal digital communication. Its training dataset based on social media gives it a sense of informal language, acronyms, hashtag usage, and platform-specific linguistic forms common in the digital environments where harassment is often found.

XLNet, proposed by Yang et al. \cite{yang2019xlnet}, provides a BERT-based architecture option. While replaced with masked language modeling, XLNet employs ``generalized autoregressive pretraining'' augmented with a permutation-based approach for bidirectional context understanding. Using this architecture, XLNet is able to connect far-text features and understand contextual signals that other language models will miss, a feature that is useful in the identification of cyberbullying content where malicious intent is typically sentence-surpassing.

TinyBERT, proposed by Jiao et al. \cite{jiao2020tinybert}, meets these computational requirements within an end-to-end knowledge distillation framework that projects knowledge from large teacher models to compact student networks. As compared to the usual distillation frameworks that just recognize output distributions, TinyBERT adopts a two-stage learning approach: general distillation on non-task-specific data and task-specific distillation. The model possesses excellent compression ratios: It compresses parameters by 7.5× and accelerates inference by 9.4× without compromising over 96\% of the teacher model's performance on GLUE tasks. When it comes to cyberbullying detection, efficiency possessed by TinyBERT also renders it extremely suitable for real-time content moderation models where response is extremely important.

DistilBERT, proposed by Sanh et al. \cite{sanh2019distilbert}, is yet another effective method for model compression, compressing BERT by 40\% while preserving 97\% of its language comprehension abilities and delivering 60\% reduced inference time. The model utilizes a distillation loss consisting of language modeling loss, soft prediction distillation loss in the teacher's predictions, and cosine embedding loss to align student and teacher embeddings.

DeBERTa (Decoding-enhanced BERT with disentangled attention), proposed by He et al. \cite{he2021deberta}, has two main innovations: a new disentangled attention mechanism and an improved masking decoder method. The model represents each token with two distinct vector representations: content and position. During attention weight calculation, the model solves three types of interactions: content-to-content, content-to-position, and position-to-content. The disentanglement enables more accurate mapping of word interdependencies, making it useful for detecting linguistic patterns typical of cyberbullying.

Hybrid systems that contain transformer architectures with dedicated classification parts have also shown to be particularly promising for cyberbullying detection. Van Bruwaene et al. \cite{vanhee2018automatic} demonstrated that the application of transformers as feature extractors coupled with traditional machine learning classifiers for decision-making performs superb detection compared to end-to-end transformer configurations. Similarly, Zhu et al. \cite{wang2021combining} proposed an integrated framework that combined Graph Convolutional Neural Networks and Label Propagation for network node classification, where architectural fusion allows the complementary strengths of different model components to be exploited.

Chen et al. \cite{du2020novel} introduce a novel Gated Broad Learning System (GBLS) over conventional Broad Learning with the addition of gate mechanisms like GRUs and LSTMs. The gates have been established to serve as adaptive filters across the network, thereby causing stable feature representation with simultaneous suppression of weak signals. In text classification, GBLS surpassed deep learning in higher precision and training speed over deep LSTM networks. The predictive performance to computational cost ratio is favorable to GBLS especially when used in cyberbullying detection. Combining effective transformer variants such as TinyBERT and DistilBERT with GBLS architectures holds an interesting prospect toward developing accurate and computation-friendly systems for cyberbullying detection.

Sentiment analysis modules facilitate cyberbullying detection by making use of tools such as VADER (Valence Aware Dictionary and sEntiment Reasoner). Hutto and Gilbert \cite{hutto2014vader} built this model based on a domain-specific dictionary and rule-based sentiment analysis to analyze sentiment across informal social media text. VADER classifies text into four categories of affect: positive affect, negative affect, neutral, and a compound score reflecting overall emotional tone. This multi-modal approach creates affective profiles for messages that enable detection of probable cyberbullying through the registration of affective connotations that are otherwise imperceptible with transformer-generated embeddings.

With automated content moderation increasingly shaping online conversation, explainability and transparency have emerged as the priority. Explainable AI (XAI) addresses the human-interpretability crisis of autonomous tech—a fundamental concern for content moderation scenarios in which automatic choices are affecting freedom of expression rights.

Zaidan et al. \cite{zaidan2007} proposed ``annotator rationales," where human annotators label certain pieces of text responsible for making their classification decisions. They showed, in their work, that such human justification supplemented training protocols not just enhances explainability but also yields high-performing models. This dual gain challenges widespread assumptions of inevitable trade-offs between performance and interpretability.

Operating under such explainability standards, Mathew et al. \cite{mathew2021} created HateXplain—a hate speech classification dataset that combines classification labels, explanatory rationales, and targeted group information. Experiments showed that human explanation-trained models improved in performance and lowered implicit bias towards commonly targeted demographic groups, with implications that explainability mechanisms can aim for accuracy and fairness in content moderation systems simultaneously.

Interpretability assessment methods have emerged to the forefront. Methods like Local Interpretable Model-agnostic Explanations (LIME)\cite{ribeiro2016why} locally using a simple and understandable model approximate a difficult model. Such methods make model decisions transparently inspectible and are therefore crucial while building trust towards automated content moderation systems.

This work tackles such gaps by proposing a hybrid architecture that combines the contextual understanding capabilities of transformer models with the pattern recognition strengths of gated learning systems, maintaining powerful interpretability in digital environments.

\section{Proposed Methodology}

The Proposed framework integrates transformer-based language understanding with specialized pattern recognition through a hybrid architecture combining ModifiedDeBERTa components with Gated Broad Learning System (GBLS) classification.

\begin{figure}[H]
\centering
\includegraphics[width=0.9\textwidth]{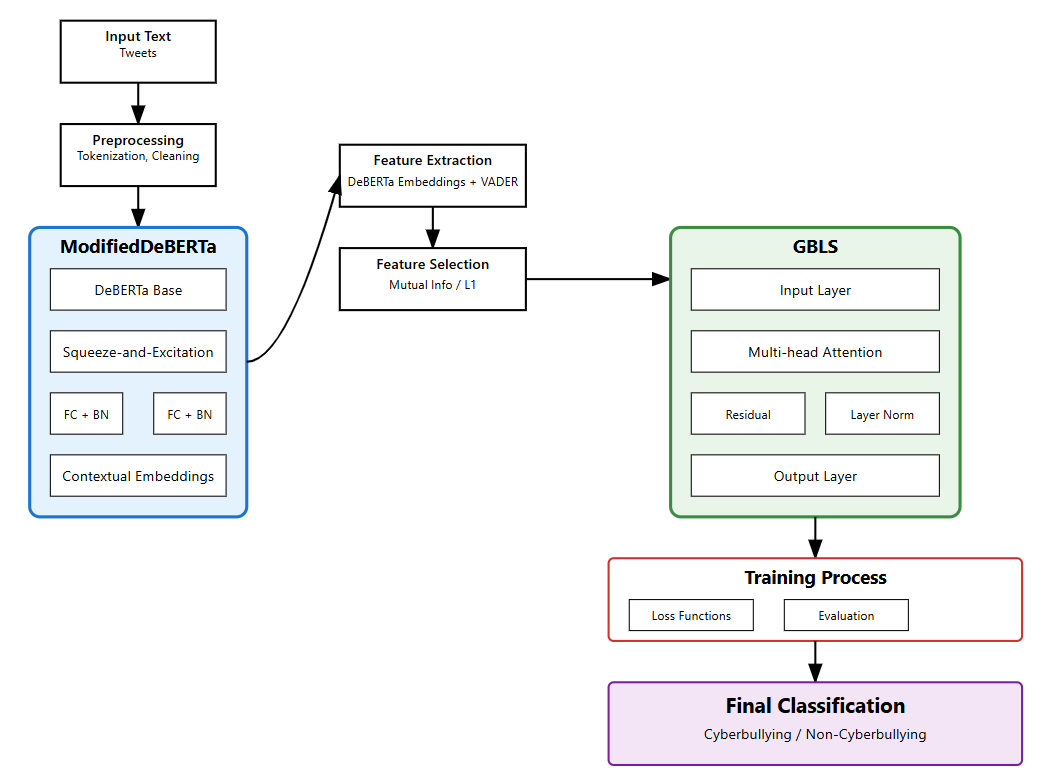}
\caption{Architecture of the proposed ModifiedDeBERTa + GBLS model.}
\label{fig:overall-architecture}
\end{figure}

As illustrated in Figure \ref{fig:overall-architecture}, the framework operates as sequential processing pipeline, beginning with input text normalization, followed by contextual representation through ModifiedDeBERTa, sentiment enhancement, feature selection, and final classification through GBLS components. Each stage contributes distinct functionality to the overall detection process.

The ModifiedDeBERTa component serves as sophisticated feature extractor, processing normalized input text to generate rich contextual representations. Built upon the pre-trained ``microsoft/deberta-v3-base'' model, this component leverages disentangled attention mechanisms where attention weights derive from content-to-content, content-to-position, and position-to-content interactions. This disentanglement enables precise modeling of word relationships, essential for detecting subtle linguistic indicators in cyberbullying content.

\begin{figure}[H]
\centering
\includegraphics[width=0.8\textwidth]{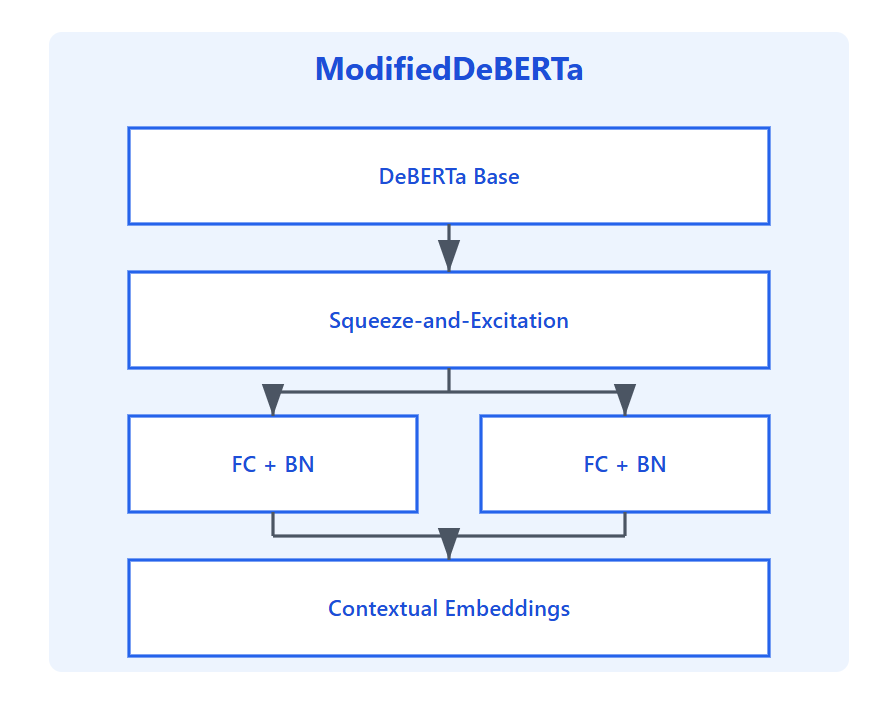}
\caption{Architecture of the ModifiedDeBERTa component showing sequential processing stages.}
\label{fig:deberta-architecture}
\end{figure}

Several architectural enhancements distinguish the ModifiedDeBERTa implementation from standard applications, as illustrated in Figure \ref{fig:deberta-architecture}. A Squeeze-and-Excitation block adaptively recalibrates channel-wise feature responses, first compressing the feature map through global pooling operations, then performing excitation through fully-connected layers with bottleneck design, followed by sigmoid activation. This mechanism allows emphasis of informative features while suppressing less useful attributes. Dimensional reduction occurs through two parallel fully-connected layers with batch normalization, progressively reducing dimensionality from 768 to 384 and subsequently to 192 dimensions. This reduction focuses representations on most salient attributes while reducing computational requirements in subsequent processing stages.

As shown in Figure \ref{fig:deberta-architecture}, the contextual embeddings generated by ModifiedDeBERTa undergo augmentation through integration with VADER sentiment scores, creating multimodal representation that captures both semantic content and affective dimensions. This integration particularly strengthens detection of implicit cyberbullying where harmful intent is conveyed through emotional undertones rather than explicitly offensive language. This augmented representation undergoes feature selection through either Mutual Information (quantifying feature-classification relationships) or L1 regularization (naturally suppressing less informative features), identifying and retaining the most discriminative signals while eliminating redundant or irrelevant information.

\begin{figure}[H]
\centering
\includegraphics[width=0.8\textwidth]{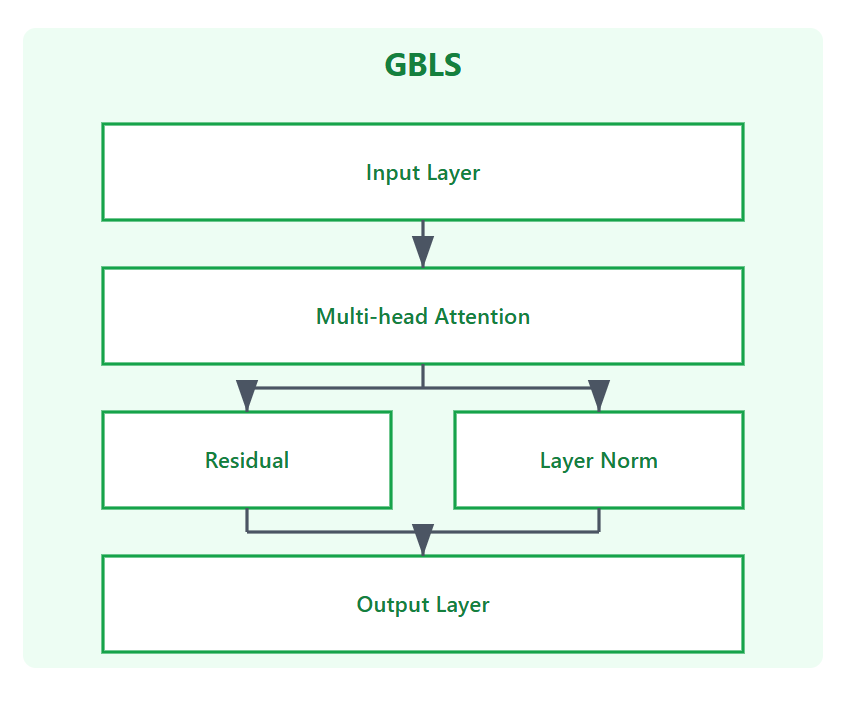}
\caption{Architecture of the GBLS classifier component.}
\label{fig:gbls-architecture}
\end{figure}

The GBLS component processes these selected features through sophisticated architectural structure designed for effective pattern recognition and classification, as illustrated in Figure \ref{fig:gbls-architecture}. The input layer receives selected features and projects them into hidden space with 216 dimensions, concentrating on relevant attributes while reducing computational complexity. A multi-head attention mechanism with 8 parallel attention heads allows the model to focus simultaneously on different feature aspects, capturing diverse patterns potentially indicative of cyberbullying. The gating mechanism also enhance detection performance by adaptively weighing the contribution of various features according to their relevance to the classification task, essentially eliminating noise while enhancing discriminative signals.

The architecture takes advantage of more than one mechanism to achieve stable and efficient learning performance. The residual connections or short-cutting paths traverse all the layers' inputs to their respective outputs, creating express paths for the information as well as gradient flow through the deep network. Shortcuts are effective in solving vanishing gradient issues where the powerful signals can be attenuated upon backpropagation. Following every critical processing stage, layer normalization keeps activation values inside a consistent range, avoiding signal sudden swings and ensuring learning stability, and convergence velocity. The last layer provides scalar value which is put through sigmoid activation function to generate probability scores indicating cyberbullying content probabilities.

The architecture has strong regularization mechanisms preinstalled against overfitting, such as dropout (random neuron disabling during training), batch normalization (keeping internal activations in within-reasonable distributional parameters), layer normalization (layer-wise outputs normalization), and L2 regularization (punishing high parametric complexity). Furthermore, highway connections—residual channels with information flow through intermediate layers—help to avoid vanishing gradient issues while enhancing training efficiency.

Algorithm \ref{alg:deberta-gbls} provides procedural description of the DeBERTa + GBLS approach, outlining the two primary stages (feature extraction using modified DeBERTa and classification using GBLS) alongside training and inference procedures.

% Add these packages to your preamble:
% \usepackage{algorithm}
% \usepackage{algorithmic}

\begin{algorithm}
\caption{Modified DeBERTa + GBLS for Cyberbullying Detection}
\label{alg:deberta-gbls}
\begin{algorithmic}[1]
\REQUIRE Training data $D = \{(X_i, y_i)\}_{i=1}^N$, Pre-trained DeBERTa model, GBLS hyperparameters
\ENSURE Fine-tuned Enhanced DeBERTa + GBLS model
\STATE \textbf{Stage 1: Fine-tune Enhanced DeBERTa}
\STATE Initialize DeBERTa with pre-trained weights
\STATE Add SE block, normalization layers, and residual connections
\FOR{each epoch}
   \FOR{each batch $(X_b, y_b)$ in $D$}
       \STATE Process inputs through DeBERTa, SE block, normalization and dense layers
       \STATE Compute total loss: $\mathcal{L} = \mathcal{L}_{cls} + 0.5 \cdot \mathcal{L}_{con}$
       \STATE Update model parameters
   \ENDFOR
   \STATE Evaluate and check early stopping criteria
\ENDFOR
\STATE \textbf{Stage 2: Extract features and train GBLS}
\STATE Initialize GBLS with specified hyperparameters
\FOR{each text input $X_i$ in training set}
   \STATE Extract and combine DeBERTa and sentiment features: $F_i = [F_i; S_i]$
\ENDFOR
\STATE Train GBLS on feature set and labels
\STATE \textbf{Inference Procedure:}
\STATE \textbf{Function} Predict(Text input $X$):
   \STATE \quad Extract and combine features
   \STATE \quad Generate GBLS prediction with optimal threshold
   \STATE \quad \textbf{return} $\hat{y}$
\end{algorithmic}
\end{algorithm}

\section{Experimental Setup}
This section describes the experimental configuration used to evaluate the proposed cyberbullying detection framework. Details are provided on the datasets employed, implementation specifications, evaluation metrics, and baseline models used for comparative assessment.

\subsection{Datasets}
Four English-language cyberbullying detection datasets were used: HateXplain, SOSNet, and two Mendeley datasets. The HateXplain dataset \cite{mathew2021} contains 20,148 Twitter and Gab tweets with comprehensive annotations. The SOSNet dataset \cite{wang2020sosnet} provides 48,000 tweets with detailed labeling structures.

The Mendeley datasets \cite{ejaz2024comprehensive} consist of two datasets—one with about 61,000 records and an enlarged dataset of about 223,000 records—collected from several social media platforms. In contrast to prior datasets centered on a single platform like Twitter, the datasets contain content from different platforms, making possible deeper insights into cyberbullying across diverse internet environments.

\subsection{Experimental Configuration}
To ensure fair comparisons, consistent testing conditions were maintained across all experiments. All models underwent evaluation on identical hardware (NVIDIA L40S GPU with 48GB memory via Lightning AI platform) using uniform software environments: PyTorch 1.13, Transformers 4.26.0, and scikit-learn 1.0.2 for models.

Six baseline models were used as comparison points to assess the suggested method: HateBERT \cite{caselli2021hatebert}, BERT variant, trained exclusively from abusive Reddit data; XLM-T \cite{barbieri2022xlmt}, transferring XLM-RoBERTa from Twitter data training; XLNet \cite{yang2019xlnet}, using generalized autoregressive pre-training with permutation-based framework for contextual understanding; BiLSTM \cite{hochreiter1997long}, bidirectional LSTM sequence processing architecture; CNN-GRU \cite{cho2014learning,lecun1998gradient}, a combination of convolutional neural networks and gated recurrent units; and BiLSTM-Attention \cite{bahdanau2014neural}, a combination of bidirectional LSTM and attention mechanisms. These baselines are disparate architecture paradigms of proven effectiveness at identifying objectionable content.

To compare the performance of various transformer models in cyberbullying detection, three variants of BERT were experimented: TinyBERT + GBLS, DistilBERT + GBLS, and DeBERTa + GBLS. In comparative analysis, the relative performance of DeBERTa with respect to DistilBERT and TinyBERT are tracked as base models to see which variant of transformer offers improved detection performance along with the GBLS module.

\subsection{Evaluation Metrics}
Performance of all models was compared through an extensive set of measurement measures to ensure vigilant assessment of detection capabilities. Five baseline metrics for classification were used: Accuracy (Acc.), which calculates the overall proportion of correct classifications; Precision (Pre.), computing the proportion of positive predictions with correct labels out of all positive predictions; Recall, computing the proportion of correct identification of actual positive cases; F1-score, computing the harmonic mean of the recall and precision to achieve balanced performance measurement; and Area Under the Curve (AUC), which computes the model's capability to discriminate between classes at various threshold levels.

Moreover, to refer to possible bias concerns in cyberbullying detection systems, Generalized Mean of Bias (GMB) scores were acquired through three test scenarios. GMB-Subgroup (GMB-Sub.) assesses bias in particular demographic or identity subgroups so that different populations of users are treated equally. GMB-Background Positive Subgroup Negative (GMB-BPSN) tests for performance when background examples are positive and subgroup examples are negative so that patterns of systematic discrimination can be confirmed. GMB-Background Negative Subgroup Positive (GMB-BNSP) addresses the contrary scenario when background cases are negative and subgroup cases are positive, with complete bias evaluation coverage.

\section{Performance Analysis on Benchmark Datasets}

This section presents detailed performance comparisons across four established benchmark datasets, demonstrating the effectiveness of the proposed approach against existing state-of-the-art methods and comprehensive baseline comparisons.
\subsection{Performance Evaluation on Individual Datasets}
This subsection presents the comprehensive performance evaluation of the proposed model across four distinct datasets, demonstrating its effectiveness and generalizability in hate speech detection tasks.

Table \ref{tab:hatexplain-results} shows relative performance on the HateXplain dataset in which the introduced DeBERTa + GBLS model recorded 79.3\% accuracy, 0.781 F1, and 0.863 ROC AUC—marked improvements over even the best baseline (HateBERT) by 3.2 percentage points in accuracy and 2.8 points in F1. DistilBERT + GBLS attained comparable performance of 76.8\% accuracy and 0.754 F1 score, whereas TinyBERT + GBLS attained 74.2\% accuracy and 0.731 F1 score, both outperforming their corresponding single-transformer baselines.

\begin{table}[H]
\centering
\caption{Performance comparison on the HateXplain dataset}
\label{tab:hatexplain-results}
\small % Reduce font size
\begin{tabular}{|c|l|c|c|c|c|c|c|}
\hline
& & \multicolumn{3}{c|}{\textbf{Performance}} & \multicolumn{3}{c|}{\textbf{Bias}} \\
\cline{3-8}
\textbf{S.No.} & \textbf{Model} & \textbf{Accuracy} & \textbf{F1} & \textbf{AUC} & \textbf{GMB-Sub.} & \textbf{GMB-BPSN} & \textbf{GMB-BNSP} \\
\hline
1 & BiLSTM & 0.643 & 0.625 & 0.801 & 0.672 & 0.641 & 0.683 \\
\hline
2 & CNN-GRU & 0.651 & 0.638 & 0.812 & 0.695 & 0.658 & 0.701 \\
\hline
3 & BiLSTM-Attention & 0.658 & 0.645 & 0.818 & 0.704 & 0.671 & 0.715 \\
\hline
4 & HateBERT & 0.761 & 0.753 & 0.858 & 0.822 & 0.760 & 0.778 \\
\hline
5 & XLM-T & 0.756 & 0.748 & 0.855 & 0.815 & 0.751 & 0.771 \\
\hline
6 & XLNet & 0.751 & 0.743 & 0.852 & 0.811 & 0.747 & 0.769 \\
\hline
7 & TinyBERT + GBLS & 0.742 & 0.731 & 0.849 & 0.812 & 0.758 & 0.771 \\
\hline
8 & DistilBERT + GBLS & 0.768 & 0.754 & 0.861 & 0.835 & 0.781 & 0.798 \\
\hline
9 & \textbf{DeBERTa + GBLS} & \textbf{0.793} & \textbf{0.781} & \textbf{0.863} & \textbf{0.866} & \textbf{0.848} & \textbf{0.863} \\
\hline
\end{tabular}

\end{table}

The table also displays bias measurements in terms of Generalized Mean of Bias (GMB) scores over three test settings: GMB-Subgroup (GMB-Sub.), GMB-Background Positive Subgroup Negative (GMB-BPSN), and GMB-Background Negative Subgroup Positive (GMB-BNSP). The reported models show considerable gains over these fairness metrics, with DeBERTa + GBLS scoring 0.866, 0.848, and 0.863 respectively, compared to HateBERT by 4.4, 8.8, and 8.5 percentage points.

The neural baseline models implemented for comprehensive comparison show performance between traditional feature-based approaches and transformer models. BiLSTM implementation achieved 64.3\% accuracy, while CNN-GRU reached 65.1\%, and BiLSTM with attention mechanisms attained 65.8\%.

On the SOSNet dataset, the proposed approach achieved good performance, as shown in Table \ref{tab:sosnet-results}. The DeBERTa + GBLS approach achieved 95.41\% accuracy and 0.9526 F1 score, outperforming the strongest comparative model (HateBERT) with gains of 2.13 percentage points in accuracy and 1.95 points in F1 score. DistilBERT + GBLS achieved 94.89\% accuracy and 0.9471 F1 score, while TinyBERT + GBLS reached 93.76\% accuracy and 0.9358 F1 score, both substantially outperforming their respective standalone baselines.

\begin{table}[H]
\centering
\caption{Performance comparison on the SOSNet dataset}
\label{tab:sosnet-results}
\begin{tabular}{|c|l|c|c|}
\hline
\textbf{S.No.} & \textbf{Model} & \textbf{Accuracy} & \textbf{F1 Score} \\
\hline
1 & BiLSTM & 0.8956 & 0.8962 \\
\hline
2 & CNN-GRU & 0.9034 & 0.9041 \\
\hline
3 & BiLSTM-Attention & 0.9089 & 0.9095 \\
\hline
4 & HateBERT & 0.9328 & 0.9331 \\
\hline
5 & XLM-T & 0.9287 & 0.9291 \\
\hline
6 & XLNet & 0.9231 & 0.9236 \\
\hline
7 & TinyBERT + GBLS & 0.9376 & 0.9358 \\
\hline
8 & DistilBERT + GBLS & 0.9489 & 0.9471 \\
\hline
9 & \textbf{DeBERTa + GBLS} & \textbf{0.9541} & \textbf{0.9526} \\
\hline
\end{tabular}
\end{table}

The Mendeley datasets provided additional validation of the proposed approaches. On the Mendeley-I dataset (Table \ref{tab:mendeley1-results}), the DeBERTa + GBLS model reached 91.37\% accuracy, 91.38\% F1 score, and 96.85\% ROC AUC, outperforming the strongest baseline by margins of 3.07\%, 3.08\%, and 1.42\% respectively. DistilBERT + GBLS achieved 89.45\% accuracy and 89.52\% F1 score, while TinyBERT + GBLS reached 87.23\% accuracy and 87.31\% F1 score.

\begin{table}[H]
\centering
\caption{Performance comparison on Mendeley-I Dataset}
\label{tab:mendeley1-results}
\footnotesize % Make font smaller
\begin{tabular}{|c|l|c|c|c|c|c|}
\hline
\textbf{S.No.} & \textbf{Model} & \textbf{Accuracy} & \textbf{Precision} & \textbf{Recall} & \textbf{F1 Score} & \textbf{ROC AUC} \\
\hline
1 & HateBERT & 0.8830 & 0.8835 & 0.8830 & 0.8830 & 0.9543 \\
\hline
2 & XLM-T & 0.8815 & 0.8823 & 0.8815 & 0.8814 & 0.9528 \\
\hline
3 & XLNet & 0.8649 & 0.8649 & 0.8649 & 0.8649 & 0.9316 \\
\hline
4 & BiLSTM & 0.8542 & 0.8456 & 0.8639 & 0.8547 & 0.9234 \\
\hline
5 & CNN-GRU & 0.8634 & 0.8567 & 0.8712 & 0.8639 & 0.9298 \\
\hline
6 & BiLSTM-Attention & 0.8681 & 0.8612 & 0.8756 & 0.8683 & 0.9327 \\
\hline
7 & TinyBERT + GBLS & 0.8723 & 0.8658 & 0.8794 & 0.8731 & 0.9412 \\
\hline
8 & DistilBERT + GBLS & 0.8945 & 0.8879 & 0.9024 & 0.8952 & 0.9567 \\
\hline
9 & \textbf{DeBERTa + GBLS} & \textbf{0.9137} & \textbf{0.9065} & \textbf{0.9213} & \textbf{0.9138} & \textbf{0.9685} \\
\hline
\end{tabular}
\end{table}

Similar patterns emerged on the Mendeley-II dataset, as shown in Table \ref{tab:mendeley2-results}, where the DeBERTa + GBLS model achieved 94.67\% accuracy, 94.73\% F1 score, and 98.23\% ROC AUC. DistilBERT + GBLS achieved 92.34\% accuracy and 92.41\% F1 score, while TinyBERT + GBLS reached 90.12\% accuracy and 90.18\% F1 score.

\begin{table}[H]
\centering
\caption{Performance comparison on Mendeley-II Dataset}
\label{tab:mendeley2-results}
\footnotesize % Make font slightly smaller than small
\begin{tabular}{|c|l|c|c|c|c|c|}
\hline
\textbf{S.No.} & \textbf{Model} & \textbf{Accuracy} & \textbf{Precision} & \textbf{Recall} & \textbf{F1 Score} & \textbf{ROC AUC} \\
\hline
1 & HateBERT & 0.9288 & 0.9301 & 0.9288 & 0.9287 & 0.9782 \\
\hline
2 & XLM-T & 0.9223 & 0.9231 & 0.9223 & 0.9223 & 0.9749 \\
\hline
3 & XLNet & 0.8914 & 0.8816 & 0.8815 & 0.8856 & 0.9428 \\
\hline
4 & BiLSTM & 0.8823 & 0.8739 & 0.8915 & 0.8826 & 0.9456 \\
\hline
5 & CNN-GRU & 0.8956 & 0.8878 & 0.9041 & 0.8959 & 0.9523 \\
\hline
6 & BiLSTM-Attention & 0.9012 & 0.8934 & 0.9096 & 0.9014 & 0.9567 \\
\hline
7 & TinyBERT + GBLS & 0.9012 & 0.8956 & 0.9074 & 0.9018 & 0.9598 \\
\hline
8 & DistilBERT + GBLS & 0.9234 & 0.9178 & 0.9301 & 0.9241 & 0.9734 \\
\hline
9 & \textbf{DeBERTa + GBLS} & \textbf{0.9467} & \textbf{0.9371} & \textbf{0.9574} & \textbf{0.9473} & \textbf{0.9823} \\
\hline
\end{tabular}
\end{table}

\subsection{Comparative Analysis with State-of-the-Art Methods}

This subsection presents a detailed evaluation of the proposed DeBERTa + GBLS based model across two benchmark English datasets, comparing its performance against established baselines for each dataset.
\begin{table}[H]
\centering
\caption{Performance comparison with State-of-the-art Model}
\label{tab:cross-dataset-comparison}
\begin{tabular}{|c|l|c|c|c|c|}
\hline
& & \multicolumn{2}{c|}{\textbf{HateXplain}} & \multicolumn{2}{c|}{\textbf{SOSNet}} \\
\cline{3-6}
\textbf{S. No.} & \textbf{Model} & \textbf{Accuracy} & \textbf{F1 Score} & \textbf{Accuracy} & \textbf{F1 Score} \\
\hline
1 & BERT-HXp \cite{mathew2021}  & 0.698 & 0.687 & - & - \\
\hline
2 & SBERT+SVM \cite{wang2020sosnet} & - & - & 0.9267 & 0.9272 \\
\hline
3 & DeBERTa + GBLS & 0.793 & 0.781 & 0.9541 & 0.9526 \\
\hline
4 & \textbf{Improvement} & \textbf{+9.5\%} & \textbf{+9.4\%} & \textbf{+2.7\%} & \textbf{+2.5\%} \\
\hline
\end{tabular}
\\[0.5ex]
\footnotesize
Note: Improvements calculated as percentage point differences.
\end{table}

Table \ref{tab:cross-dataset-comparison} compares the proposed DeBERTa + GBLS approach against the best performing baseline models from the original dataset publications. The results show performance improvements on both datasets, with larger gains observed on HateXplain (9.5\% accuracy improvement) compared to SOSNet (2.7\% accuracy improvement). The varying improvement margins reflect the different characteristics and complexity levels of the respective datasets.

\section{Explainability Framework}

The explainability module in the proposed cyberbullying detection framework is designed to improve the transparency, interpretability, and trustworthiness of automated decisions. As content moderation systems are increasingly deployed in socially sensitive environments, it becomes essential to provide meaningful explanations for model predictions. This paper integrates three complementary interpretability techniques: token-level attribution analysis, instance-level explanations using LIME, and confidence calibration diagnostics.

\subsection{Token-level Attribution Analysis}
Token-level attribution analysis identifies individual words or phrases that have the strongest influence on the model's predictions. Gradient-based attribution methods, such as Integrated Gradients~\cite{sundararajan2017axiomatic}, are applied to compute the contribution of each token to the final classification result.
This analysis showed that the model attends to both explicit offensive terms and implicit indicators of bias or hostility. Table~\ref{tab:token-attribution} presents examples of tokens with high attribution scores from the Mendeley-I dataset, organized by category and average contribution values.
\begin{table}[h]
\centering
\caption{Token Attribution Analysis for Cyberbullying Detection (Mendeley-I Dataset)}
\label{tab:token-attribution}
\begin{tabular}{|c|l|l|c|c|}
\hline
\textbf{S.No.} & \textbf{Category} & \textbf{Word/Phrase} & \textbf{Avg. Attribution} & \textbf{Frequency} \\
\hline
\multirow{3}{*}{1} & \multirow{3}{*}{Explicit Offensive} & fucking & 0.67 & High \\
\cline{3-5}
& & stupid  & 0.52 & High \\
\cline{3-5}
& & idiot   & 0.48 & Medium \\
\hline
\multirow{2}{*}{2} & \multirow{2}{*}{Implicit Bias} & people like you & 0.39 & Medium \\
\cline{3-5}
& & your kind       & 0.36 & Low \\
\hline
3 & Context Modifiers & seriously       & 0.31 & Medium \\
\hline
\multirow{2}{*}{4} & \multirow{2}{*}{Targeting Pronouns} & you             & 0.35 & Very High \\
\cline{3-5}
& & your            & 0.32 & Very High \\
\hline
\multirow{2}{*}{5} & \multirow{2}{*}{Negative Sentiment} & hate            & 0.45 & Medium \\
\cline{3-5}
& & disgusting      & 0.42 & Low \\
\hline
\end{tabular}
\end{table}
The analysis reveals that the model effectively identifies both explicit offensive language and subtle contextual indicators of cyberbullying behavior.
\subsection{LIME-based Local Interpretations}

To explain individual predictions, the framework employs Local Interpretable Model-Agnostic Explanations (LIME)~\cite{ribeiro2016why}. LIME approximates the behavior of a complex model using an interpretable surrogate model within the local neighborhood of a specific input, highlighting the tokens that contribute most to the final prediction.

\begin{figure}[H]
\centering
\includegraphics[width=0.75\textwidth]{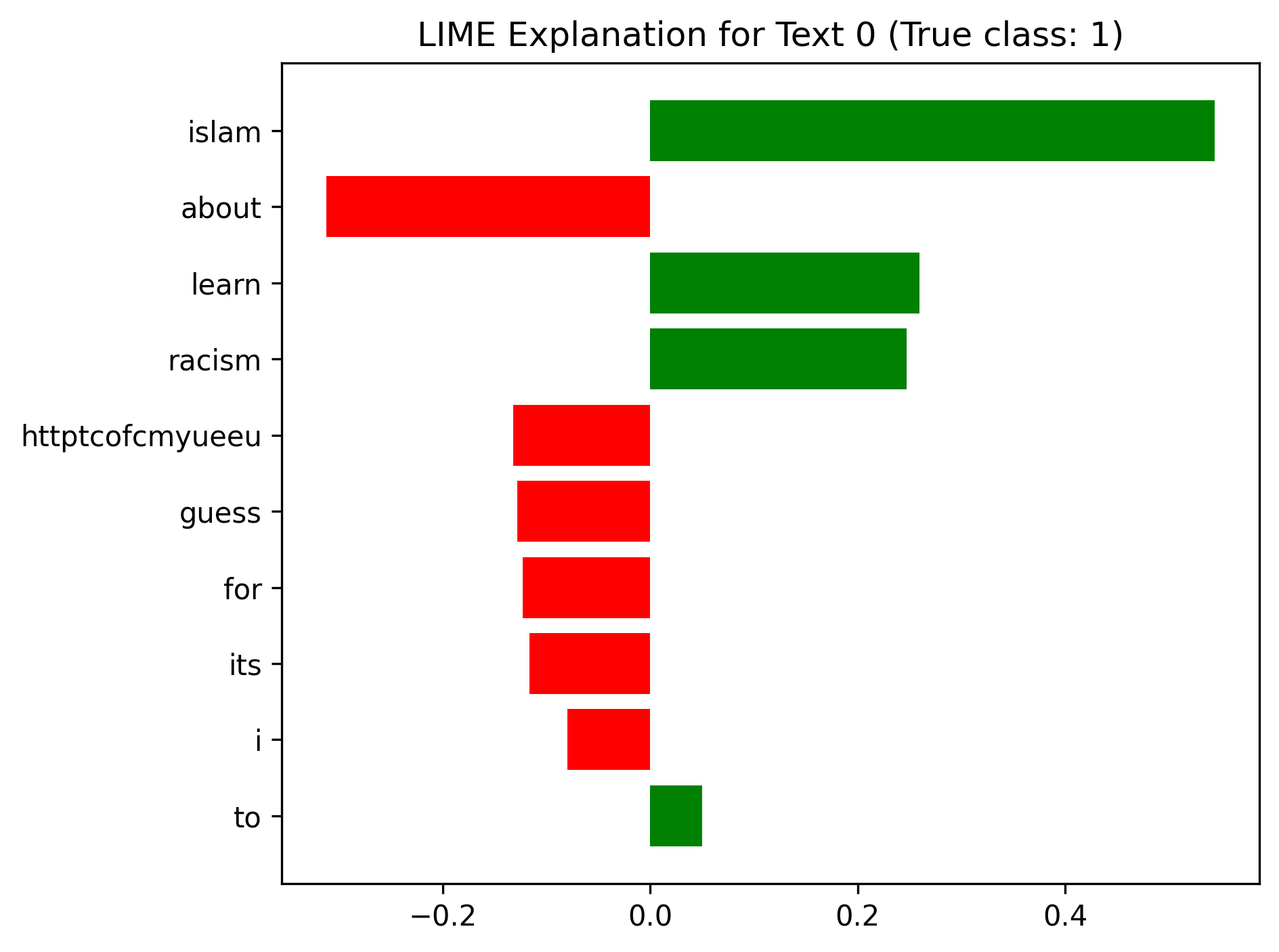}
\caption{LIME explanation showing word-level contributions to a cyberbullying prediction.}
\label{fig:lime-example}
\end{figure}

An example LIME explanation for a cyberbullying prediction from the Mendeley-I dataset is illustrated in Figure~\ref{fig:lime-example}. In this case, the analysis shows that words such as \textit{Islam}, \textit{learn}, and \textit{racism} positively contributed to the harmful classification, while neutral terms like \textit{about} reduced the prediction score.

These local explanations are particularly valuable for understanding model behavior in borderline cases and for identifying potential misclassifications.

\subsection{Confidence Calibration Analysis}

In addition to attribution-based explanations, the framework evaluates model confidence through calibration analysis. Neural models are known to exhibit overconfidence, which can be problematic in high-stakes applications. A well-calibrated model produces probability scores that reflect the true likelihood of correct classification~\cite{guo2017calibration}.

\begin{figure}[h!]
\centering
\includegraphics[width=0.75\textwidth]{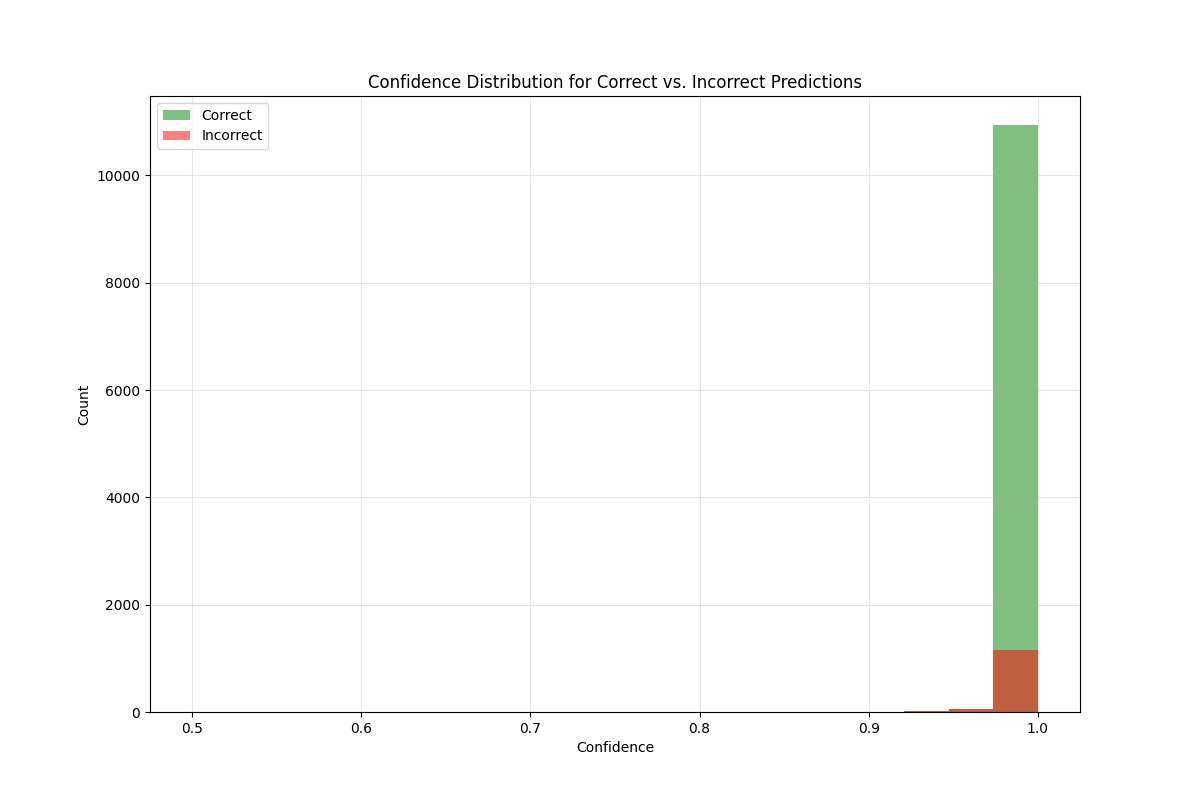}
\caption{Distribution of model confidence scores on the Mendeley-I test set.}
\label{fig:confidence-distribution}
\end{figure}

The distribution of confidence scores for predictions made on the Mendeley-I test set is shown in Figure~\ref{fig:confidence-distribution}. Most predictions are concentrated near high-confidence thresholds, although some errors still occur at extreme confidence levels, indicating opportunities for improvement.

Confidence calibration supports human-in-the-loop moderation strategies by identifying cases where automatic decisions can be trusted and when human review is advisable.

\subsection{Model Ablation Study}

The ablation study for the detection framework examined three key augmentation components: (1) the Squeeze-and-Excitation block, (2) VADER sentiment integration, and (3) feature selection mechanisms, with GBLS serving as the essential classifier component throughout all configurations. The Mendeley-I dataset was selected for ablation analysis due to its substantial size enabling robust component evaluation and its binary classification structure facilitating clearer interpretation of individual component contributions. Table \ref{tab:english_ablation} presents the systematic addition of enhancement components evaluated on the Mendeley-I dataset.

\begin{table}[H]
\centering
\caption{Ablation Study Results for English Detection Framework (Mendeley-I Dataset)}
\label{tab:english_ablation}
\footnotesize % Make font smaller
\setlength{\tabcolsep}{2pt} % Reduce column spacing
\begin{tabular}{|c|l|c|c|c|c|c|}
\hline
\textbf{S.No.} & \textbf{Configuration} & \textbf{Accuracy} & \textbf{Precision} & \textbf{Recall} & \textbf{F1-Score} & \textbf{AUC} \\
\hline
1 & DeBERTa + GBLS (baseline) & 0.9078 & 0.9008 & 0.9151 & 0.9079 & 0.9638 \\
\hline
2 & DeBERTa + GBLS + SE Block & 0.9089 & 0.9019 & 0.9162 & 0.9090 & 0.9651 \\
\hline
3 & DeBERTa + GBLS + SE + VADER & 0.9112 & 0.9042 & 0.9185 & 0.9113 & 0.9669 \\
\hline
4 & \textbf{DeBERTa + GBLS + SE + VADER + FS} & \textbf{0.9137} & \textbf{0.9065} & \textbf{0.9213} & \textbf{0.9138} & \textbf{0.9685} \\
\hline
\multicolumn{7}{|c|}{\textit{Component Contribution Analysis}} \\
\hline
5 & SE Block Contribution & +0.0011 & +0.0011 & +0.0011 & +0.0011 & +0.0013 \\
\hline
6 & VADER Sentiment Contribution & +0.0023 & +0.0023 & +0.0023 & +0.0023 & +0.0018 \\
\hline
7 & Feature Selection Contribution & +0.0025 & +0.0023 & +0.0028 & +0.0025 & +0.0016 \\
\hline
\multicolumn{7}{|c|}{\textit{Feature Selection Method Comparison}} \\
\hline
8 & Mutual Information & 0.9137 & 0.9065 & 0.9213 & 0.9138 & 0.9685 \\
\hline
9 & L1 Regularization & 0.9124 & 0.9054 & 0.9197 & 0.9125 & 0.9672 \\
\hline
10 & No Feature Selection & 0.9112 & 0.9042 & 0.9185 & 0.9113 & 0.9669 \\
\hline
\end{tabular}
\end{table}

The ablation results reveal that each augmentation component contributes meaningfully to overall performance when combined with the essential GBLS classifier. The Squeeze-and-Excitation block provides a 0.11 percentage point improvement in accuracy, demonstrating its effectiveness in recalibrating feature importance. VADER sentiment integration contributes 0.23 percentage points, highlighting the value of affective information for cyberbullying detection. Feature selection adds another 0.25 percentage points, confirming the benefit of focusing on the most discriminative features.

Feature selection method comparison indicates that Mutual Information outperforms L1 regularization (91.37\% vs 91.24\% accuracy), suggesting that information-theoretic approaches are more effective for identifying relevant features in cyberbullying contexts.

\subsection{Model Failure Cases}
Misclassified cases by the detection model indicate persistent patterns of error that demonstrate intrinsic difficulties in automated cyberbullying detection. Analysis of such failure cases sheds light on model limitations and areas for improvement. Representative samples of hard cases for the DeBERTa + GBLS model are shown in Table \ref{tab:english_failure_cases}.

\begin{table}[h]
\centering
\caption{Failure case analysis for English cyberbullying detection}
\label{tab:english_failure_cases}
\small
\setlength{\tabcolsep}{4pt}
\begin{tabular}{|c|p{6cm}|c|p{1cm}|p{4cm}|}
\hline
\textbf{S.No.} & \textbf{Text} & \textbf{Truth} & \textbf{Pred.} & \textbf{Analysis} \\
\hline
1 & ``Some people are just \textcolor{orange}{naturally better} at certain things. It's not \textcolor{orange}{discrimination}, it's reality." & \begin{tabular}[c]{@{}c@{}}Cyber\\bully\end{tabular} & \begin{tabular}[c]{@{}c@{}}Non\\bully\end{tabular} & Subtle implicit bias masked as objective statement. Model failed to detect veiled discriminatory intent. \\
\hline
2 & ``Your argument is completely \textcolor{red}{invalid} and shows your \textcolor{red}{ignorance} on this topic." & \begin{tabular}[c]{@{}c@{}}Non\\bully\end{tabular} & \begin{tabular}[c]{@{}c@{}}Cyber\\bully\end{tabular} & Strong critique mistaken for personal attack. Model over-sensitive to negative language in academic contexts. \\
\hline
3 & ``Thanks for \textcolor{red}{ruining my day} with your \textcolor{green}{brilliant} insights." & \begin{tabular}[c]{@{}c@{}}Cyber\\bully\end{tabular} & \begin{tabular}[c]{@{}c@{}}Non\\bully\end{tabular} & Sarcastic cyberbullying with positive words. Model struggled with ironic intent detection. \\
\hline
4 & ``People like you make me \textcolor{red}{sick}. This \textcolor{red}{behavior} needs to \textcolor{red}{stop immediately}." & \begin{tabular}[c]{@{}c@{}}Non\\bully\end{tabular} & \begin{tabular}[c]{@{}c@{}}Cyber\\bully\end{tabular} & Counter-speech against harmful behavior misclassified due to presence of aggressive language. \\
\hline
\end{tabular}
\end{table}

\textbf{Color Coding Legend:}
\begin{itemize}
\item \textcolor{red}{Red highlighting}: Negative/aggressive language that may trigger false positives.
\item \textcolor{orange}{Orange highlighting}: Subtle bias indicators that evade detection.
\item \textcolor{green}{Green highlighting}: Positive words used sarcastically or ironically.
\end{itemize}

The English model failure analysis reveals four primary challenge categories:

\textbf{Implicit Bias Detection:} The model struggles with finer discrimination that bypasses outright offensive language. Offensive stereotypes created by apparently objective assertions typically go undetected. Typically, this happens when it applies neutral-sounding assertions with discriminatory overtones.

\textbf{Context-Dependent Criticism:} Scholarly or professional rejection with very derogatory language is at times mistakenly labeled as cyberbullying. This pattern is triggered by very negative adjectives employed within valid criticism, especially in scholarly disagreement contexts.

\textbf{Sarcasm and Irony:} Sarcastic cyberbullying using positive words for negative ends is especially problematic. Difficulty for the model in detecting ironic subversion of meaning causes classification errors. This involves cases in which what seems like positive language is used to communicate hurtful intent.

\textbf{Counter-speech Misclassification:} Replies to opposing or denying harmful conduct have been referred to as cyberbullying because of the harshness of their wording. They vary from strong words employed for the purpose of discouraging harmful actions instead of continuing with them.
\section{Conclusion and Future Work}

This research provides an end-to-end detection model of cyberbullying which appropriately integrates the contextual understanding ability of transformer models with the pattern learning ability of gated broad learning systems. The proposed ModifiedDeBERTa + GBLS architecture presented here is a testament to sustained performance improvements for a wide range of benchmarking data sets and obtains significant accuracy improvements over state-of-the-art baselines while having good fairness metrics along with offering meaningful explainability. The combination of VADER sentiment analysis and feature selection mechanisms enhances the model to identify more subtle forms of cyberbullying that are emotionally underpinned in content instead of overtly offensive content. The four English test datasets confirm the effectiveness of the approach, and the explainability framework meets stringent transparency requirements in automatic content moderation systems. The ablation studies conducted clearly support that each architectural element individually makes significant contributions to end performance, with the Squeeze-and-Excitation block, sentiment unification, and feature selection making incremental accuracy gains. Yet, failure case evaluation identifies ongoing issues with implicit bias detection, sarcastic content detection, and differentiation from true criticism, where current methods are still lacking.

Future research would involve solving the particular weaknesses of this work. First, efforts of calibration methods to deal with model overconfidence, especially in those instances in which the model is confident with erroneous predictions. The same can be done through temperature scaling, Platt scaling, or advanced uncertainty quantification techniques to get more accurate confidence estimations. Second, targeted improvements for the failure cases identified in this study, including enhanced sarcasm detection through pragmatic inference models, improved implicit bias recognition using subtle linguistic cue analysis, better context-dependent criticism handling through discourse-aware architectures, and refined counter-speech identification mechanisms that can distinguish protective responses from harmful content.


\begin{thebibliography}{99}

\bibitem{patchin2023cyberbullying}
J. W. Patchin and S. Hinduja, "Cyberbullying trends among teens: A nationally representative sample," Cyberbullying Research Center, 2023.

\bibitem{kowalski2014psychological}
R. M. Kowalski, G. W. Giumetti, A. N. Schroeder, and M. R. Lattanner, "Bullying in the digital age: A critical review and meta-analysis of cyberbullying research among youth," \textit{Psychological Bulletin}, vol. 140, no. 4, pp. 1073--1137, 2014.

\bibitem{schmidt2017}
A. Schmidt and M. Wiegand, "A survey on hate speech detection using natural language processing," in \textit{Proceedings of the Fifth International Workshop on Natural Language Processing for Social Media}, 2017, pp. 1--10.

\bibitem{davidson2017}
T. Davidson, D. Warmsley, M. Macy, and I. Weber, "Automated hate speech detection and the problem of offensive language," in \textit{Proceedings of the 11th International AAAI Conference on Web and Social Media}, 2017, pp. 512--515.

\bibitem{badjatiya2017}
P. Badjatiya, S. Gupta, M. Gupta, and V. Varma, "Deep learning for hate speech detection in tweets," in \textit{Proceedings of the 26th International Conference on World Wide Web Companion}, 2017, pp. 759--760.

\bibitem{hochreiter1997long}
S. Hochreiter and J. Schmidhuber, "Long short-term memory," \textit{Neural Computation}, vol. 9, no. 8, pp. 1735--1780, 1997.

\bibitem{cho2014learning}
K. Cho, B. Van Merriënboer, C. Gulcehre, D. Bahdanau, F. Bougares, H. Schwenk, and Y. Bengio, "Learning phrase representations using RNN encoder-decoder for statistical machine translation," in \textit{Proceedings of the 2014 Conference on Empirical Methods in Natural Language Processing}, 2014, pp. 1724--1734.

\bibitem{caselli2021hatebert}
T. Caselli, V. Basile, J. Mitrović, and M. Granitzer, "HateBERT: Retraining BERT for abusive language detection in English," in \textit{Proceedings of the 5th Workshop on Online Abuse and Harms}, 2021, pp. 17--25.

\bibitem{barbieri2020tweeteval}
F. Barbieri, J. Camacho-Collados, L. Espinosa Anke, and L. Neves, "TweetEval: Unified benchmark and comparative evaluation for tweet classification," in \textit{Findings of the Association for Computational Linguistics: EMNLP 2020}, 2020, pp. 1644--1650.

\bibitem{yang2019xlnet}
Z. Yang, Z. Dai, Y. Yang, J. Carbonell, R. Salakhutdinov, and Q. V. Le, "XLNet: Generalized autoregressive pretraining for language understanding," in \textit{Advances in Neural Information Processing Systems}, 2019, pp. 5753--5763.

\bibitem{he2021deberta}
P. He, X. Liu, J. Gao, and W. Chen, "DeBERTa: Decoding-enhanced BERT with disentangled attention," in \textit{Proceedings of the 9th International Conference on Learning Representations}, 2021.

\bibitem{jiao2020tinybert}
X. Jiao, Y. Yin, L. Shang, X. Jiang, X. Chen, L. Li, F. Wang, and Q. Liu, "TinyBERT: Distilling BERT for natural language understanding," in \textit{Findings of the Association for Computational Linguistics: EMNLP 2020}, 2020, pp. 4163--4174.

\bibitem{sanh2019distilbert}
V. Sanh, L. Debut, J. Chaumond, and T. Wolf, "DistilBERT, a distilled version of BERT: smaller, faster, cheaper and lighter," in \textit{Proceedings of the 7th Workshop on Energy Efficient Machine Learning and Cognitive Computing}, 2019.

\bibitem{hutto2014vader}
C. J. Hutto and E. Gilbert, "VADER: A parsimonious rule-based model for sentiment analysis of social media text," in \textit{Proceedings of the Eighth International AAAI Conference on Weblogs and Social Media}, 2014, pp. 216--225.

\bibitem{du2020novel}
J. Du, C.-M. Vong, and C. L. P. Chen, "Novel efficient RNN and LSTM-like architectures: Recurrent and gated broad learning systems and their applications for text classification," \textit{IEEE Transactions on Cybernetics}, vol. 51, no. 3, pp. 1586--1597, 2020.

\bibitem{vanhee2018automatic}
C. Van Hee, G. Jacobs, C. Emmery, B. Desmet, E. Lefever, B. Verhoeven, G. De Pauw, W. Daelemans, and V. Hoste, "Automatic detection of cyberbullying in social media text," \textit{PLoS One}, vol. 13, no. 10, pp. e0203794, 2018.

\bibitem{wang2021combining}
H. Wang and J. Leskovec, "Combining graph convolutional neural networks and label propagation," \textit{ACM Transactions on Information Systems}, vol. 40, no. 4, pp. 1--27, 2021.

\bibitem{zaidan2007}
O. Zaidan, J. Eisner, and C. Piatko, "Using 'annotator rationales' to improve machine learning for text categorization," in \textit{Human Language Technologies 2007: The Conference of the North American Chapter of the Association for Computational Linguistics}, 2007, pp. 260--267.

\bibitem{mathew2021}
B. Mathew, P. Saha, S. M. Yimam, C. Biemann, P. Goyal, and A. Mukherjee, "HateXplain: A benchmark dataset for explainable hate speech detection," in \textit{Proceedings of the 35th AAAI Conference on Artificial Intelligence}, 2021, pp. 14867--14875.

\bibitem{wang2020sosnet}
J. Wang, K. Fu, and C. T. Lu, "SOSNet: A graph convolutional network approach to fine-grained cyberbullying detection," in \textit{2020 IEEE International Conference on Big Data}, 2020, pp. 1699--1708.

\bibitem{ejaz2024comprehensive}
N. Ejaz, S. Choudhury, and F. Razi, "A comprehensive dataset for automated cyberbullying detection," Mendeley Data V2, doi: 10.17632/wmx9jj2htd.2, 2024.

\bibitem{ribeiro2016why}
M. T. Ribeiro, S. Singh, and C. Guestrin, "``Why should I trust you?'': Explaining the predictions of any classifier," in \textit{Proceedings of the 22nd ACM SIGKDD International Conference on Knowledge Discovery and Data Mining}, 2016, pp. 1135--1144.

\bibitem{guo2017calibration}
C. Guo, G. Pleiss, Y. Sun, and K. Q. Weinberger, "On calibration of modern neural networks," in \textit{International Conference on Machine Learning}, 2017, pp. 1321--1330.

\bibitem{sundararajan2017axiomatic}
M. Sundararajan, A. Taly, and Q. Yan, "Axiomatic attribution for deep networks," in \textit{International Conference on Machine Learning}, 2017, pp. 3319--3328.

\bibitem{barbieri2022xlmt}
F. Barbieri, L. E. Anke, and J. Camacho-Collados, "XLM-T: Multilingual Language Models in Twitter for Sentiment Analysis and Beyond," in \textit{Proceedings of the Language Resources and Evaluation Conference}, 2022, pp. 258--266.

\bibitem{lecun1998gradient}
Y. LeCun, L. Bottou, Y. Bengio, and P. Haffner, "Gradient-based learning applied to document recognition," \textit{Proceedings of the IEEE}, vol. 86, no. 11, pp. 2278--2324, 1998.

\bibitem{bahdanau2014neural}
D. Bahdanau, K. Cho, and Y. Bengio, "Neural machine translation by jointly learning to align and translate," arXiv preprint arXiv:1409.0473, 2014.

\end{thebibliography}
\end{document}